\documentclass[11pt,letterpaper]{article}
\usepackage{naaclhlt2015}
\usepackage{times}
\usepackage{latexsym}

\usepackage[hyphens]{url}

\usepackage{latexsym}
\usepackage{multirow}
\usepackage{multicol}
\usepackage{booktabs}
\usepackage{color}
\usepackage{graphicx} 
\usepackage{amsmath}

\usepackage{pifont}
\newcommand{\cmark}{\ding{51}}%
\newcommand{\xmark}{\ding{55}}%

\usepackage{multirow}
\title{Echoes of Persuasion: \\ The Effect of Euphony in Persuasive Communication}

\author{Marco Guerini \\
  Trento RISE\\
  Povo, I-38100 Trento\\
  {\tt \footnotesize{marco.guerini@trentorise.eu}}\\\And
  G\"ozde \"Ozbal \\
  FBK-Irst\\
  Povo, I-38100 Trento\\
  {\tt \footnotesize{gozbalde@gmail.com}} \\\And
  Carlo Strapparava \\
  FBK-Irst\\
  Povo, I-38100 Trento\\
  {\tt \footnotesize{strappa@fbk.eu}} \\}
  
\date{}

\begin{document}
\maketitle
\begin{abstract}
While the effect of various lexical, syntactic, semantic and stylistic features have been addressed in persuasive language from a computational point of view, the persuasive effect of phonetics has received little attention. By modeling a notion of euphony 
and analyzing four datasets comprising persuasive and non-persuasive sentences in different domains (political speeches, movie quotes, slogans and tweets), we explore the impact of sounds on different forms of persuasiveness. We conduct a series of analyses and prediction experiments within and across datasets. Our results highlight the positive role of phonetic devices on persuasion.

\end{abstract}

\section{Hocus Pocus}


Historically, in human sciences, several definitions of persuasion have been proposed -- see for example \cite{Toulmin:58,walton:96,Chaiken1980,cial93,Petty1986}. Most of them have a common core addressing:
\emph{methodologies aiming to change the mental state of the receiver by means of communication in view of a possible action to be performed by her/him.} \cite{perelman:olbrechts:69,moul02}.

These methodologies might take into account the overall structure of a text such as the ordering of the arguments or simply single word choices. For a successful text both of them are often required.
The focus of persuasion may vary according to the goal of the communication and it can take different forms according to the domain: from memorability (e.g., making people remember a statement or a product) to diffusion (e.g., making people pass on a content in social networks by sharing it), from behavioral change (e.g., political communication) to influencing purchasing decisions (e.g., slogans to convince people to try or buy a product) -- see for example \cite{heath2007made}.
\textcolor{black}{While many techniques such as resorting to expert opinion, utilizing the framing effect, emotive language or exaggeration can be used to obtain such persuasive effects, we devote this study to explore particular techniques pertaining to euphony.}

Euphony refers to the inherent pleasantness of the sounds of words, phrases and sentences, and it is utilized to achieve pleasant, rhythmical and harmonious effects. The idea that the pleasantness of the sounds in a sentence can foster its effectiveness is rooted in our culture, and is connected to the concepts of rhythm and music. 
The fact that language and music interact in our brain has been shown by localizing low-level syntactic processes of music and language in the temporal lobe \cite{sammler2013co}. It has also been shown that changes in the cardiovascular and respiratory systems can be induced by music -- specifically tempo, rhythm, melodic structure \cite{bernardi2006cardiovascular}.
The importance of euphony has its roots also in ancient human psychology.  
As Julian Jaynes suggests \cite{jaynes2000origin}, poetry used to be divine knowledge. 
It was the sound and tenor of authorization and it commanded where plain prose could only ask. A paradigmatic example of this conception is the act of casting a spell. 
Spells (incantations) are special linguistic objects that are meant not only to change how people think or behave but they are also so powerful that they can -- allegedly  -- change reality. Spells are often very euphonic (and meaningless) sentences, e.g. ``Hocus Pocus".


Various psycholinguistic studies addressed the effects of phonetics on the audience in different aspects such as memorability \cite{wales,doi:10.1080/10926488.2013.797728} or more specifically advertisement \cite{leech,soundAdvice}. There are also computational studies that address the problem of recognizing persuasive sentences according to various syntactic, lexical and semantic features \cite{danescu2012you,tan2014effect}. However, to the best of our knowledge, the direct impact of phonetic elements on persuasiveness has not been explored in computational settings yet. 

In this paper, we fill in this gap by conducting a series of analyses and prediction experiments on four datasets representing different 
aspects of persuasive language to evaluate the importance of a set of phonetic devices (i.e. rhyme, alliteration, homogeneity and plosives) on various forms of persuasiveness. 
Our experiments show that phonetic features play an important role in the detection of persuasiveness and encode a notion of ``melodious language" that operates both within and across datasets.


\section{Related Work}\label{relatedWork}


In the following, we first revise some 
NLP studies addressing linguistic features of successful communication. Then, we summarize a selection of studies devoted to the effects of phonetics on persuasion. 

\subsection{NLP studies on persuasion}
\newcite{virality} focus on a particular form of persuasion by using New York Times articles to examine the relationship between 
virality (i.e., the tendency of a content to be circulated on the Web) and emotions evoked by the content. They conduct semi-automated sentiment analysis to quantify the affectivity 
and emotionality of each article. Results suggest a strong relationship between affect and virality, in this case 
measured as the count of how many people emailed each article. As suggested by the authors, this metric represents a form of ``narrowcasting", as opposed to other ``broadcasting" actions such as sharing on Twitter.   

Another line of research investigates the impact of various textual features on audience reactions. The work by \newcite{marco:carlo:gozde:ICWSM-11} correlates several viral 
phenomena with the wording of a post,  while \newcite{guerini2012linguistic} show that features such as the readability level of an abstract influence the number of downloads, 
bookmarking and citations.

A particular approach to content virality is presented by \newcite{simmons2011memes}, 
who explore the impact of different types of modification on memes spreading from one person to another.

\newcite{danescu2012you} measure a different ingredient of persuasion by analyzing the features of a movie quote that make it ``memorable". They compile a corpus consisting of memorable and non-memorable movie quote pairs and conduct a detailed analysis to investigate the lexical and syntactic differences between these pairs.

\newcite{louis2013makes} focus on influential science articles in newspapers by considering characteristics such as readability, description vividness, use of unusual words 
and affective content. High quality articles (NYT articles appearing in ``The Best American Science Writing" anthology) are compared against typical NYT articles.  

\newcite{borghol2012untold} investigate how differences in textual description affect the spread of content-controlled videos.
\newcite{lakkaraju2013s} focus on the act of resubmissions (i.e., content that is submitted multiple times with multiple titles to multiple different communities) to understand the extent to which each factor influences the success of a content. \newcite{tan2014effect} consider how content spreads in an on-line community by pinpointing the effect of wording in terms of content informativeness, generality and affect.  
\newcite{althoff2014ask} develop a model that can predict the success of requests for a free pizza gifted from the Reddit community. 
The authors consider high-level textual features such as politeness, reciprocity, narrative and gratitude.


\subsection{Studies on the effects of phonetics} 
\newcite{doi:10.1080/10926488.2013.797728} states that alliteration and rhyme can be considered as attention-seeking devices as they enhance emphasis. The author also suggests that they are useful for acceptability and long-term retention of original expressions, decrypting their meanings, indicating informality, and breaking the ice between an audience and a speaker. Therefore, these devices are commonly used in original metaphorical and metonymical compounds.

According to \newcite{leech}, phonetic devices such as rhyme and alliteration are systematically exploited by advertisers to achieve 
memorability. Similarly, \newcite{wales} underlines the effectiveness of alliteration and rhyme on emphasis and memorability of an expression.

The relation between the usage of plosives (i.e., consonants in which the vocal tract is blocked so that all airflow ceases, such as ``p", ``t" or ``k") and memorability has also been investigated. According to the study carried out by \newcite{soundAdvice} brand names starting with plosive sounds are recalled and recognized more than the ones starting with other sounds. \newcite{ozbal:brand} carry out an analysis of brand names and discover that plosives are very commonly used.

\newcite{danescu2012you}, whom we previously mentioned, carry out an auxiliary analysis and observe the differences in letter and sound distribution (e.g. usage of labials or front vowels, back sounds, coordinating conjunctions) of memorable and non-memorable quotes. 

\newcite{BrainSup} propose a phonetic scorer for creative sentence generation such that generated sentences can contain various phonetic features including alliteration, rhyme and plosive sounds. The authors evaluate the proposed model on automatic slogan generation. In a more recent work \cite{P14-2058}, they enforce the existence of these features in the sentences that are automatically generated for second language learning to introduce hooks to echoic memory. 

\section{Phonetic Scorer}
\label{PhoneticScorer}
For the design of the phonetic features, we were mostly inspired by the work of \newcite{BrainSup}, who built and used three phonetic scorers for creative sentence generation. Similarly to this work, all the phonetic features that we used are based on the phonetic representation of English words of the Carnegie Mellon University pronouncing dictionary\footnote{The CMU pronunciation dictionary is freely available at \url{http://www.speech.cs.cmu.edu/cgi-bin/cmudict}. We have used version 0.7a in our implementation.}. We selected four classes of phonetic devices, namely plosives, alliteration, rhyme and homogeneity, which can easily be modeled by observing the distribution of specific classes of phonemes within the sentence.
   The \emph{plosive} score is calculated as the ratio of the number of plosive sounds in a sentence to the overall number of phonemes. For both \emph{alliteration} and \emph{rhyme} scorers, we provide a na\"{\i}ve implementation that does not consider stresses or syllables, but only counts the number of repeated sounds at the beginning or end of words in the sentence. The alliteration score is calculated as the number of repeated phonetic prefixes in a sentence normalized by the total number of phonemes. Similarly, the \emph{rhyme} score is calculated as the ratio of the number of repeated phonetic endings in a sentence to the total number of phonemes. Lastly, the homogeneity scorer simply calculates the degree of homogeneity in terms of phonemes used in a sentence independently from their positions. If we let $d_{\textrm{ph}}$ be the count of distinct phonemes and $t_{\textrm{ph}}$ be the total count of phonemes in a sentence, then the homogeneity score is calculated as $1 - (d_{\textrm{ph}} / t_{\textrm{ph}})$.

\section{Dataset}
\label{DS}

In this section, we describe the four datasets we used to conduct our analyses and experiments. 
As we mentioned previously, the definition of persuasion is a debated topic and it can comprise distinct strategies or facets. For this reason, we experimented with datasets where at least one ingredient is clearly in the equation.
To explore the effects of wording and euphonics on persuasion, the datasets were built in a controlled setting (topic, author, sentence length) to avoid confounding factors such as author or topic popularity, by following the procedure described in \cite{danescu2012you,tan2014effect}. In addition, these datasets comprise short texts (mostly single sentences) to focus on surface realization of persuasion, where strategic planning -- which might act as a confounding factor -- plays a minor role.
The idea of using controlled experiments (usually in an A/B test setting) to study persuasive communication can be traced back at least to Hovland et al. (1953). 
While two of these datasets (Twitter and Movies) were already available, the other two (CORPS and Slogans) were collected by following the methodology proposed in the first two as closely as possible\footnote{CORPS and Slogans datasets can be downloaded at the following link: \url{https://github.com/marcoguerini/paired_datasets_for_persuasion/}}. 

All datasets are built around the core idea of collecting pairs consisting of a persuasive sentence ($P$) and a non-persuasive counterpart ($\neg P$), where $P$ and $\neg P$ are structurally very similar and controlled for the above mentioned confounding factors. 

\textbf{Twitter}. A set of 11,404  
tweet pairs, where each pair comes from the same user (author control) and contains the same URL (topic control).
$P$ and $\neg P$ are determined based on their retweet counts  \cite{tan2014effect}. 
It is worth noting that in our experiments we were able to collect only 11,019 of such tweet pairs since some of them were deleted in the meanwhile. 

\textbf{Movie}. A set of 2,198 single-sentence memorable movie quotes ($P$) paired with non-memorable quotes ($\neg P$). For each $P$, the dataset contains a contrasting  
quote $\neg P$ from the same movie such that 
 (i) $P$ and $\neg P$ are uttered by the same speaker, (ii) $P$ and $\neg P$ have the same number of words, (iii) $\neg P$ does not occur in the IMDb list of memorable quotes and (iv) $P$ and $\neg P$ are as close as possible to each other in the script 
\cite{danescu2012you}.

\textbf{\textsc{CORPS}}. A set of 2,600 sentence pairs uttered by various politicians. We collected these pairs from CORPS, a freely available corpus of political speeches tagged with audience reactions \cite{guerini2013new}. The methodology that we used to build the pairs is very similar to \newcite{danescu2012you}: for each $P$, where $P$ is the sentence preceding an audience reaction (e.g. \texttt{APPLAUSE}, \texttt{LAUGHTER}), we selected a contrasting single-sentence $\neg P$ from the same speech. We required $\neg P$ to be close to $P$ in the speech transcription, subject to the conditions that (i) $P$ and $\neg P$ are uttered by the same speaker - which is trivial since these are monologues, where a single speaker is addressing the audience - (ii) $P$ and $\neg P$ have the same number of words, and (iii) $\neg P$ is 5 to 15 sentences away from $P$. This last condition had to be imposed since, differently from movie quotes, we do not have the evidence of which fragment of the speech exactly provoked the audience reaction (i.e. it could be the combination of more than one sentence).  

\textbf{Slogan}. A set of 1,533 slogans taken from online resources 
paired with non-slogans that are similar in content. 
We collected the non-slogans from the subset of the New York Times articles in English GigaWord -- 5th Edition -- released by Linguistic Data Consortium
(LDC)\footnote{\url{http://www.ldc.upenn.edu/Catalog/catalogEntry.jsp?catalogId=LDC2011T07}}. For each slogan, we picked the 
most similar sentence in the New York Times articles having the same length and the highest LSA similarity \cite{Deerwester90indexingby} with 
the slogan. The LSA similarity approach that we used to collect the non-slogans is very similar to the approach used by 
\newcite{louis2013makes} to collect the non-persuasive counterparts of successful news articles. 

In Table \ref{tab:dataset_charact}, we sum up the criteria used in the construction of each dataset. As can be observed from the table, each dataset satisfies at least two of the three criteria described above. 
\begin{table} [!htb] 	
 \centering
 \footnotesize	
		\begin{tabular}{lcccrr} 		
			\toprule
			& \multicolumn{3}{c}{Criterion} & \multicolumn{2}{c}{Length}\\
 DATASET &   Author &  Length &  Topic & \multicolumn{1}{c}{$P$} & \multicolumn{1}{c}{$\neg P$}  \\
 \cmidrule(r){1-1}  \cmidrule(r){2-4} \cmidrule(r){5-6}
\textsc{CORPS}  &\cmark &  \cmark & \xmark & 14.0 & 14.0 \\
Movie &\cmark & \cmark & \xmark  & 9.7 & 9.7\\
Slogan  & \xmark & \cmark & \cmark & 5.0  & 5.0 \\
Twitter  &\cmark & \xmark & \cmark & 16.2 & 15.4 \\
\bottomrule 
		\end{tabular} 	
	\caption{Criteria used in the construction of each dataset and average token length of persuasive and non-persuasive pairs} 	
	\label{tab:dataset_charact} 
\end{table}  
In the last two columns of the table, we also provide the average token length of the persuasive and non-persuasive sentences in each dataset. Finally, in Table~\ref{tab:dataset_examples} we provide examples of euphonic and persuasive sentences for each dataset together with their phonetic scores.

\begin{table*} [!htb] 	
 \centering
 \scriptsize	
		\begin{tabular}{llrrrr} 		
		 
			\toprule
 Dataset & Example & Rhyme & Alliteration & Plosive & Homogeneity\\
\midrule			
\multirow{2}{*}{CORPS} & I think we can do better and I think we must do better. & 0.789 & 0.737 & 0.342 & 0.631\\
& It will be waged with determination and it will be waged until we win. & 0.566 & 0.679 & 0.189 & 0.736\\
\midrule
\multirow{3}{*}{Movie}& The night time is the right time. & 0.818 & 0.545 & 0.181 & 0.636\\
& Beautiful.... beautiful butterfly... & 0.667 & 0.708 & 0.250 & 0.583\\
& Dog eat dog, brother. & 0.533 & 0.533 & 0.400 & 0.400\\
\midrule

\multirow{3}{*}{Slogan} &Different Stores, Different Stories. & 0.621 & 0.896 & 0.207 & 0.690\\
& Why ask why? Try Bud Dry & 0.625 & 0.625 & 0.312 & 0.437\\
&Live, Love, Life. & 0.818 & 0.909 & 0.0 & 0.636\\
\midrule

\multirow{2}{*}{Twitter} & A Nerd in Need is a Nerd indeed. & 0.636 & 0.727 & 0.227 & 0.681\\
& Easter cupcake baking!! & 0.0 & 0.0 & 0.412 & 0.470\\

            \bottomrule 
            
		\end{tabular} 	
		
	\caption{Euphonic examples of persuasive sentences from each dataset, along with their phonetic scores.} 	
	\label{tab:dataset_examples} 
\end{table*}


\section{Data Analysis} \label{sec:quantitative}
\begin{table*}[t]
   \centering
   \scriptsize
    \begin{tabular}{lllllllll}
        \toprule
        
& \multicolumn{2}{c}{Rhyme}&\multicolumn{2}{c}{Alliteration} & \multicolumn{2}{c}{Plosive} & \multicolumn{2}{c}{Homogeneity}\\
Dataset & \multicolumn{1}{c}{\emph{$\mu$}} & \multicolumn{1}{c}{\emph{$\sigma$}} 
& \multicolumn{1}{c}{\emph{$\mu$}} & \multicolumn{1}{c}{\emph{$\sigma$}} 
& \multicolumn{1}{c}{\emph{$\mu$}} & \multicolumn{1}{c}{\emph{$\sigma$}} 
& \multicolumn{1}{c}{\emph{$\mu$}} & \multicolumn{1}{c}{\emph{$\sigma$}}\\
\cmidrule(r){1-1} \cmidrule(l){2-3} \cmidrule(l){4-5} \cmidrule(l){6-7} \cmidrule(l){8-9}
CORPS$_{\neg P}$& 0.233 & 0.143 & 0.208 & 0.142 & 0.187 & 0.058 & 0.603 & 0.173\\
CORPS$_P$ & 0.245$\dagger$ & 0.152 & 0.223** & 0.154 & 0.194*** & 0.060 & 0.588** & 0.179\\
\cmidrule(r){1-1} \cmidrule(l){2-3} \cmidrule(l){4-5} \cmidrule(l){6-7} \cmidrule(l){8-9}
Movie$_{\neg P}$ & 0.196 & 0.143 & 0.167 & 0.142 & 0.191 & 0.073 & 0.485 & 0.155 \\
Movie$_P$ & 0.214* & 0.165 & 0.196*** & 0.164 & 0.185$\dagger$ & 0.067 & 0.526*** & 0.164\\
\cmidrule(r){1-1} \cmidrule(l){2-3} \cmidrule(l){4-5} \cmidrule(l){6-7} \cmidrule(l){8-9}
Slogan$_{\neg P}$& 0.071 & 0.111 & 0.047 & 0.092 & 0.204 & 0.098 & 0.343 & 0.163\\
Slogan$_P$ & 0.140*** & 0.194 & 0.123*** & 0.185 & 0.189*** & 0.098 & 0.366*** & 0.156\\
\cmidrule(r){1-1} \cmidrule(l){2-3} \cmidrule(l){4-5} \cmidrule(l){6-7} \cmidrule(l){8-9}
Twitter$_{\neg P}$& 0.204 & 0.116 & 0.180 & 0.114 & 0.188 & 0.058 & 0.617 & 0.134 \\
Twitter$_P$ & 0.216*** & 0.121 & 0.193*** & 0.120 & 0.185** & 0.055  & 0.636*** & 0.128\\

        \bottomrule
    \end{tabular}
    \caption{Average phonetic scores for our datasets - ***, $p<.001$; **, $p<.01$; *, $p<.05$; $\dagger$, not significant} 	
	\label{tab:dataset_scores} 
\end{table*}

To provide a first insight on the data, in Table \ref{tab:dataset_scores} we report the average phonetic scores for each data set (Mann-Whitney U Test is used for statistical significance between $P$ and $\neg P$ samples, with Bonferroni correction to ameliorate issues with multiple comparisons). The results are partially in line with our expectations of the euphony phenomena being more relevant in the persuasive sentences across the datasets. 

As can be observed from the table, the average rhyme scores are higher in persuasive sentences and the difference is highly significant for Slogan and Twitter ($p<.001$), slightly significant for Movie quotes ($p<.05$), but not significant for CORPS. The average alliteration scores are again higher in persuasive sentences and all the differences are highly significant in all datasets (apart from CORPS with $p<.01$). Plosives seem not to correlate well with our intuition of persuasiveness and euphony: either there is no significance (movie quotes) or the averages of euphonic scores are higher in the non-persuasive sentences (the difference is highly significant in slogans, and significant in Twitter). The only dataset that meets our expectation is CORPS with a highly significant difference in favor of persuasive sentences. Finally, the average homogeneity scores are significantly ($p<.001$) higher in persuasive sentences in all datasets except CORPS, where the scores of non-persuasive sentences are significantly higher ($p<.01$) than persuasive ones. 

Without going into details of cross-dataset comparisons we would like to note that CORPS seems a very peculiar dataset in terms of average scores, as compared to the others. In terms of rhyme and alliteration, the average scores of non-persuasive sentences ($\neg P$) in CORPS are always higher than the persuasive sentences ($P$) in the other datasets ($p<.001$ in all cases), while for homogeneity the same holds apart from Twitter. 
These results may derive from the fact that a political speech is a carefully crafted text -- aimed at influencing the audience -- in its entirety, so also ``non-persuasive" sentences in CORPS are on average more persuasive than in other datasets. 



As a next step, we conducted another analysis on the distribution of ``extreme cases", i.e. sentences that have a very high phonetic score at least in one feature. This analysis derives from the intuition that a euphonic sentence might be recognized as such by humans only if its phonetic scores are above a certain threshold. In fact, sound repetition in a sentence may occur by chance, as in ``I saw \textbf{the} knife in \textbf{the} drawer", and the longer the sentence is, the higher the probability that phonetic scores will be non-zero even in absence of a euphonic effect. Therefore, the average scores for each phonetic device, as reported in Table \ref{tab:dataset_scores}, are only partially informative. 

Given this premise, 
to evaluate the ``persuasive power" of the phonetic devices taken into account, we compare them in terms of empirical Complementary Cumulative Distribution Functions (CCDFs) of the persuasive/non-persuasive pairs in various datasets. 
These functions are commonly used to analyze online social networks in terms of growth in size and activity (see for example \cite{ahn2007,jiang2010,leskovec2008dynamics}) and also for measuring content diffusion, e.g. the number of retweets of a given content \cite{kwak2010}. Here, we use CCDFs to account for the probability $P$ that the score of a phonetic device \emph{d} will be greater than $n$ indicating it with $\hat{F}_{d}(n)$.
For example, the
probability of having a text with more than .75 rhyme score is indicated with $\hat{F}_{rh}(.75) = \mathrm{P}
(\mathrm{\#rhyme}>.75)$. 
To assess whether the CCDFs of the several types of texts we take into account show significant differences, 
we use the Kolmogorov-Smirnov goodness-of-fit test, which specifically targets cumulative distribution functions. 
In particular, for each phonetic device and dataset, we use a two-tailed Kolmogorov-Smirnov test (again with Bonferroni correction) to test whether the number of examples above the threshold is higher in the persuasive sentences than in their non-persuasive counterparts for that device.

 
\begin{table}[t] 	
	\centering	
	\scriptsize
		\begin{tabular}{lllll} 
		\toprule
             \emph{Dataset} & $\hat{F}_{rh}(t)$ & $\hat{F}_{al}(t)$ & $\hat{F}_{pl}(t)$ & $\hat{F}_{ho}(t)$\\
             \midrule
            CORPS$_{\neg P}$ & 0.025 & 0.012 & 0.362 & 0.394\\
            CORPS$_{P}$ & 0.033\scriptsize{**} & 0.023\scriptsize{**} & 0.415\scriptsize{***} & 0.363\scriptsize{$\dagger$}\\
            \midrule
            Movie$_{\neg P}$ & 0.018 & 0.011 & 0.397 & 0.092\\
            Movie$_{P}$ & 0.041\scriptsize{***} & 0.025\scriptsize{***} & 0.363\scriptsize{$\dagger$} & 0.173\scriptsize{***}\\
             \midrule
            Slogan$_{\neg P}$ & 0.005 & 0.003 & 0.460 & 0.011\\
            Slogan$_{P}$ & 0.055\scriptsize{***} & 0.043\scriptsize{***} & 0.410\scriptsize{***} & 0.018\scriptsize{***}\\
             \midrule
            Twitter$_{\neg P}$ & 0.006 & 0.003 & 0.385 & 0.377\\
            Twitter$_{P}$ & 0.012\scriptsize{***} & 0.008\scriptsize{***} & 0.364\scriptsize{**} & 0.449\scriptsize{***}\\
            \bottomrule 
		\end{tabular} 		
	\caption{Probability of examples above threshold, - ***, $p<.001$; **, $p<.01$; *, $p<.05$; $\dagger$, not significant}
	\label{tab:ccdf} 
\end{table} 

Since we do not have a theoretical way to define such thresholds, we resort to empirically define them by using a specific dataset of euphonic sentences. Even if it might seem reasonable to consider poems as paradigmatic examples of ``euphonic" writing, we discard them as the phonetic devices used in poems  
 may span across sentences. Instead, we resort to tongue twisters as a gold reference of how a euphonic sentence should be. Accordingly, we collected a set of 534 tongue twisters from various online resources. Then, for each phonetic index we defined our thresholds as the average of the phonetic scores in this data, in particular: $t_{rh} = 0.55$ for rhyme, $t_{al} = 0.58$ for alliteration, $t_{pl} = 0.20$ for plosives and $t_{ho} = 0.68$ for homogeneity. 

In Table~\ref{tab:ccdf}, we report the results of our CCDF analysis. 
After analyzing the ``extreme cases", where euphony is granted, we see that the trends found in Table \ref{tab:dataset_scores} on the correlation between persuasiveness and euphony are confirmed and strengthened. The number of persuasive sentences with a rhyme score above threshold is 30\% more than the non-persuasive ones in CORPS, while the difference is 90\% in Twitter\footnote{In the following the ratios are computed on the real values while Table~\ref{tab:ccdf} presents the rounded values.}. The ratio of persuasive sentences above threshold to non-persuasive ones is very high in movies and slogans (more than 2 and 10 respectively).  All results are either highly significant or significant. For comparison, in Table \ref{tab:dataset_scores} these differences are not significant for CORPS and only slightly significant ($p<.05$) for movies.
Concerning alliteration, there are 85\% more cases above threshold in the persuasive sentences of CORPS than the non-persuasive ones. For movie quotes and Twitter, the persuasive sentences above threshold are more than two times as many as the non-persuasive ones, while the ratio is more than 13 for slogans. All results are either highly significant or significant in line with the results of Table \ref{tab:dataset_scores}. Instead, for plosive scores we observe a negative or no correlation with persuasiveness, the only exception being CORPS.
Regarding homogeneity, for CORPS the difference between persuasive and non-persuasive sentences is not significant (in Table \ref{tab:dataset_scores} it was significantly in favor of non-persuasive sentences), while for the other datasets there is a highly significant difference in favor of persuasive sentences (between 20\% and 80\%). As a whole, these results confirm our intuition that phonetic features play a significant role with respect to persuasiveness. In the next section we will validate this claim by means of prediction experiments.

\section{Prediction Experiments}
In this section, we describe the prediction tasks (both within and across datasets) that we carried out to investigate the impact of the phonetic features on the detection of various forms of persuasiveness. We compare three different sets of features, namely phonetic, n-grams and their combination to understand whether phonetic information can improve the performance of standard lexical approaches. 
Similarly to \newcite{danescu2012you} and \newcite{tan2014effect}, we formulate a pairwise classification problem such that given a pair $(s_{1},s_{2})$ consisting of sentences $s_{1}$ and $s_{2}$, the goal is to determine the more persuasive one (i.e., the one on the \emph{left} or \emph{right}). We can consider this as a binary classification task where for each instance (i.e., pair) the possible labels are \emph{left} or \emph{right}.

\subsection{Dataset and preprocessing}
For the prediction experiments, we used the four datasets described in Section~\ref{DS} (i.e., CORPS, Twitter, Slogan and Movie), all of which consist of a persuasive sentence $P$ and its non-persuasive counterpart ($\neg P$) labeled as either \emph{left} or \emph{right}. To make the positions of the sentences in a pair irrelevant (i.e. to provide symmetry), for each instance occurring in the original datasets (e.g., $(s_{1},s_{2})$ with label \emph{left}), we added another instance including the same sentence pair in reverse order (i.e., $(s_{2},s_{1})$ with label \emph{right}). 
As a preprocessing step, all the sentences were tokenized by using 
Stanford CoreNLP~\cite{manning-EtAl:2014:P14-5}.

\subsection{Classifier and features}
We performed a 10-fold cross-validation on each dataset and experimented with three feature sets by using a Support Vector Machine (SVM) classifier~\cite{svm}. 
We preferred SVM as a classifier due to its characteristic property to especially perform well on high-dimensional data \cite{Weichselbraun:2011:UGP:2063576.2063729}.

The first feature set consists of the phonetic features (i.e. plosive, alliteration, rhyme and homogeneity scores as detailed in Section~\ref{PhoneticScorer}). The second feature set is a standard bag of word n-grams including unigrams, bigrams and trigrams. All the non-ascii characters, punctuations and numbers were ignored. The URLs and mentions in Twitter data were replaced with tags (i.e. \emph{\_URL\_} and \emph{\_MENTION\_} respectively). In addition, for the unigram features, stop words were filtered out. 
We did not apply this filtering for bigrams and trigrams to capture longer-range usage patterns such as propositional phrases. The third feature set is simply the union of both phonetic and n-gram features.  

To find the best configuration for each dataset and feature set, we conducted a grid search over the degree of the polynomial kernel (1 or 2) and the number of features to be used (in the range between 1,000 and 20,000). Due to the low dimensionality of the phonetic feature set, feature selection was performed only for the feature sets including n-grams. The selection was performed based on the information gain of each feature.

\begin{table}[!htb]
   \centering
 \resizebox{\linewidth}{!}{
 \begin{tabular}{lccc}
        \toprule
       \emph{Dataset} & \emph{Phonetic} & \emph{N-Gram} & \emph{All} \\
        \cmidrule(r){1-1} \cmidrule(l){2-4}
        CORPS & 0.589 (-, 1) & $0.733^{***}$ (4k, 1) & $0.736^\dagger\hspace{0.82em}$ (2k, 1) \\
        Movie &  0.600 (-, 2) & $0.694^{***}$ (1k, 1) &  $0.722^{***}$ (1k, 1) \\
        Slogan & 0.700 (-, 2) & $0.826^{***}$ (3k, 1) & $0.883^{***}$ (5k, 1)\\
        Twitter & 0.563 (-, 2) & $0.732^{***}$ (5k, 1) & $0.745^{***}$ (4k, 1) \\
        \bottomrule
    \end{tabular}
    }
    \caption{Results of the within-dataset experiments.}
    \label{tab:withinDomain}
\end{table}

\begin{table*}[!htb]
   \centering
   \footnotesize
 \begin{tabular}{lccccc}
        \toprule
       \emph{Dataset} & \emph{N-Gram} & \emph{N-Gram+Rhyme} & \emph{N-Gram+Plosive} & \emph{N-Gram+Homogeneity} & \emph{N-Gram+Alliteration}\\
        \cmidrule(r){1-1} \cmidrule(l){2-6}
        
        CORPS   & 0.733   & $0.738^\dagger\hspace{0.82em}$ (3k, 1) & $0.740^\dagger\hspace{0.82em}$ (2k, 1) & $0.738^\dagger\hspace{0.82em}$ (3k, 1) & $0.738^\dagger\hspace{0.82em}$ (2k, 1) \\
        
        Movie   & 0.694   & $0.694^\dagger\hspace{0.82em}$ (1k, 1) & $0.692^\dagger\hspace{0.82em}$ (1k, 1) & $0.721^{***}$ (1k, 1) & $0.709^{**}\hspace{0.44em}$ (1k, 1) \\
         
        Slogan  & 0.826   & $0.864^{***}$ (2k, 1) & $0.824^\dagger\hspace{0.82em}$ (2k, 1) & $0.867^{***}$ (3k, 1) & $0.859^{***}$ (3k, 1) \\       
        Twitter & 0.732   & $0.740^{**}\hspace{0.44em}$ (4k, 1) & $0.733^\dagger\hspace{0.82em}$ (4k, 1) & $0.746^{***}$ (4k, 1) & $0.742^{***}$  (4k, 1) \\
        
        \bottomrule
    \end{tabular}
    \caption{Contribution of the phonetic features.}
    \label{tab:phonetic_exp}
\end{table*}

\begin{table*}[!htb]
   \centering
    \resizebox{\linewidth}{!}{
    \begin{tabular}{lcccccccccccc}
        \toprule
        & \multicolumn{12}{c}{\emph{Test}} \\
        & \multicolumn{3}{c}{CORPS}&\multicolumn{3}{c}{Twitter} & \multicolumn{3}{c}{Slogan} & \multicolumn{3}{c}{Movie}\\
       \emph{Training} & \emph{Phonetic} & \emph{N-Gram} & \emph{All} & \emph{Phonetic} & \emph{N-Gram} & \emph{All} & \emph{Phonetic} & \emph{N-Gram} & \emph{All} & \emph{Phonetic} & \emph{N-Gram} & \emph{All}\\
        \cmidrule(r){1-1} \cmidrule(l){2-4} \cmidrule(l){5-7} \cmidrule(l){8-10} \cmidrule(l){11-13}
        CORPS & - & - & - & 0.463 & 0.508 & 0.523 & 0.508 & 0.517 & 0.539 & 0.411 & 0.506 & 0.516\\
        Twitter & 0.439 & 0.494 & 0.462 & - & - & - & 0.564 & 0.531 & 0.637 & 0.596 & 0.544 & 0.589 \\
        Slogan & 0.535  & 0.512 & 0.514 & 0.535 & 0.510 & 0.539 & - & - & - & 0.532 & 0.545 & 0.588  \\
        Movie & 0.431 & 0.513 & 0.498 & 0.562  & 0.533 & 0.560  & 0.581 & 0.537 & 0.589 & - & - & -  \\
    
        \bottomrule
    \end{tabular}
    }
    \caption{Results of the cross-dataset prediction experiments optimized on the training set.}
    \label{tab:performance_crossDom_optimizedOnTr}
\end{table*}

\subsection{Within-dataset experiments}
For this set of experiments, we conducted a 10-fold cross validation on each dataset separately.
In Table~\ref{tab:withinDomain}, for each dataset listed in the first column, in the subsequent columns we report the performance of the best model obtained with 10-fold cross validation using i) only phonetic features (\emph{Phonetic}), ii) only n-grams (\emph{N-Gram}), iii) both phonetic and n-gram features (\emph{All}). As mentioned previously, for each pair $(s_{1},s_{2})$ consisting of sentences $s_{1}$ and $s_{2}$, our dataset contains another pair including the same sentences in  reverse order (i.e., $(s_{2},s_{1})$), resulting in a symmetric and balanced dataset. Therefore, classification performance is measured in terms of accuracy (i.e., the percentage of pairs of which labels were correctly predicted). For each accuracy value, we also report in parenthesis the number of features selected and the kernel degree of the corresponding model. While the kernel degree did not make a big difference in the performance, the number of selected features had an important effect on the accuracy of the models. As can be observed from these values, the best performance on all the datasets is achieved with a relatively small number of features.  

Among the values reported in the table, the ones followed by $^{***}$ are significantly different ($p<.001$) from the ones to their left, while $\dagger$ represents no significance, as calculated according to McNemar's test~\cite{McNemar47samplingIndependence}. 
For each dataset, the weakest models (i.e. the ones using only the phonetic features in all cases) are still significantly ($p<.001$) more accurate than a random baseline (accuracy = 50\%). 
As can be observed from the table, the models using only n-grams significantly outperform the ones only based on phonetic features in all datasets. However, while the phonetic features are not very strong by themselves, their combination with n-grams results in models outperforming the n-gram based models in all cases. The difference is highly significant for all datasets except CORPS, where n-grams alone are sufficient to achieve a good performance. We speculate that the kind of persuasiveness used in political speeches is more dependent on the lexical choices of the speaker and on the use of a specific set of semantically loaded words such as \emph{bless}, \emph{victory}, \emph{God} and \emph{justice} or \emph{military}. This is in line with the work of \newcite{guerini2008corps}, who built a domain specific lexicon to study the persuasive impact of words in political speeches.  

We also conducted an additional set of experiments to investigate if some phonetic features stand out among the others, and to find out the contribution and importance of each phonetic feature in isolation. 
To achieve that, for each dataset we conducted a 10-fold cross validation to obtain the best four models containing a single phonetic feature on top of n-gram features (i.e. \emph{N-Gram+Rhyme}, \emph{N-Gram+Plosive}, \emph{N-Gram+Homogeneity} and \emph{N-Gram+Alliteration}). In Table~\ref{tab:phonetic_exp}, we report the accuracy of the n-gram model and these four models for each dataset. Similarly to Table~\ref{tab:withinDomain}, for each accuracy value, we also report in parenthesis the number of features selected and the kernel degree of the corresponding model obtained with grid search. The results demonstrate that homegeneity is the most effective feature when added on top of n-grams, resulting in highly significant improvement against the basic n-gram models in three out of four datasets. Alliteration and rhyme closely follow homogeneity by yielding models that significantly outperform the n-gram models in three and two datasets respectively. Finally, the models containing plosives do not improve over the n-gram models in any of the four datasets. It is worth noting that in CORPS none of the n-gram models enriched with phonetic features improves over the basic n-gram models as in line with the results of the within-dataset experiments reported in Table \ref{tab:withinDomain}.

\subsection{Cross-dataset experiments}

After observing that the combination of phonetic and n-gram features can be effective in the within-dataset prediction experiments, we took a further step and investigated the interaction of the three feature sets across datasets. More specifically, we classified each dataset with the best models (one for each feature set) trained on the other datasets. With these experiments, we investigated the ability of phonetic features to generalize across the different lexicons of the datasets. As we discussed previously, the four datasets represent different forms of persuasiveness. In this respect, the results of the cross-dataset experiments can also be interpreted as a measure of the degree of compatibility among these kinds of persuasiveness.

In Table~\ref{tab:performance_crossDom_optimizedOnTr}, we present the results of the cross-dataset prediction experiments. For each training and test set pair, 
we report the accuracy of the best models, one for each feature set, based on cross-validation on the training set. As can be observed from the table, the figures are generally low and various domain adaptation techniques could be employed to improve the results. However, the objective of this evaluation is not to train an optimized cross-domain classifier, but to assess the potential of the feature sets to model different kinds of persuasiveness.

As expected, n-gram features show poor performance due to the lexical and stylistic differences among the datasets. In many cases, the phonetic models outperform the n-gram models, and in several cases the combination of the two feature sets deteriorates the performance of the phonetic features alone. These findings support our hypothesis that phonetic features, due to their generality, have better correlation with different forms of persuasiveness than lexical features. The experiments involving the CORPS dataset, both for training and testing, do not share this behavior. Indeed, when CORPS is used as a training or test dataset, the performance of the models is quite low (very close to or worse than the baseline in many cases) independently from the feature sets. These results suggest that the notion of persuasiveness encoded in this dataset is remarkably different from the others, as previously discussed in the data analysis in Section~\ref{sec:quantitative}. As seen in the within dataset experiments (see Table~\ref{tab:withinDomain}), CORPS is the only dataset in which the combination of lexical and phonetic features do not improve the classification accuracy. This explains the inability of the phonetic features to improve the accuracy in cross-dataset experiments when this dataset is employed.

    

\section{Conclusion}

In this paper, we focused on the impact of a set of phonetic features -- namely rhyme, alliteration, homogeneity and plosives -- on various forms of persuasiveness including memorability of slogans and movie quotes, re-tweet counts of tweets, and effectiveness of political speeches. We conducted our analysis and experiments on four datasets comprising pairs of a persuasive sentence and a non-persuasive counterpart. 

Our data analysis shows that persuasive sentences are generally euphonic. 
This finding is confirmed by the prediction experiments, in which we observed that phonetic features consistently help in the detection of persuasiveness. When combined with lexical features, they help improving classification performance on three of the four datasets that we considered. The key role played by phonetic features is further underlined by the cross-dataset experiments, in which we observed that phonetic features alone generally outperform the lexical features. 
To the best of our knowledge, this is the first systematic analysis of the impact of phonetic features on several types of persuasiveness. Our results should encourage researchers dealing with different aspects of persuasiveness to consider the inclusion of phonetic attributes in their models.


As future work, we will investigate the impact of other phonetic devices such as assonance, consonance and rhythm on persuasiveness. It would also be interesting to focus on the connection between sound symbolism and persuasiveness, and investigate how the context or domain of persuasive statements interacts with the sounds in those statements.

\vspace{0.50em}

\noindent We would like to conclude this paper with the most favorite and retweeted tweet of @NAACL2015 (the Twitter account of the conference whose proceedings comprise this paper), which is a good example of the positive effect of euphony in persuasiveness:

\begin{quote}
\center
\small
\emph{The deadline for @NAACL2015 paper submissions is approaching: \\ Remember, remember, the 4th of December!}
\end{quote}



\section*{Acknowledgments}
This work has been partially supported by the Trento RISE PerTe project.

\bibliographystyle{naaclhlt2015}
\bibliography{Persuasive}

\end{document}